\documentclass[twoside]{article}

\usepackage[preprint]{aistats2026}
\usepackage[round]{natbib}

\usepackage{bm}
\usepackage{algorithm}
\usepackage{algpseudocode}
\usepackage{hyperref}
\usepackage{url}
\usepackage{graphicx}
\usepackage{amsthm} 
\usepackage{amsmath,amssymb,thmtools}
\usepackage{xcolor}

\newtheorem{proposition}{Proposition}

\DeclareMathOperator*{\argmax}{argmax}
\DeclareMathOperator*{\argmin}{argmin}

\begin{document}

%

%


\twocolumn[

\aistatstitle{Rethinking Trust Region Bayesian Optimization in High Dimensions}

\aistatsauthor{Wei-Ting Tang \And Joel A. Paulson }

\aistatsaddress{ University of Wisconsin--Madison \And University of Wisconsin--Madison } ]

\begin{abstract}
Trust Region Bayesian Optimization (TuRBO) is an effective strategy for alleviating the curse of dimensionality in high-dimensional black-box optimization. However, inappropriate lengthscale design can cause the local Gaussian process (GP) model within the trust region to degenerate, leading to suboptimal performance in high dimensions.
In this work, we show that TuRBO’s local GP may remain either excessively complex or overly simple as the dimension $D$ and trust region side length $L$ vary. To address this issue, we propose a straightforward variant, \textbf{AdaScale-TuRBO}, which scales the GP lengthscale with both the problem dimension and trust region size, thereby preserving kernel geometry and maintaining consistent prior complexity. Empirically, we show that AdaScale-TuRBO can robustly outperform standard TuRBO and other popular high-dimensional BO methods on synthetic benchmarks and real-world trajectory planning tasks.
\end{abstract}

\section{INTRODUCTION}
\label{sec:intro}

Bayesian optimization (BO) \citep{garnett2023bayesian} is a powerful framework for optimizing expensive black-box functions where each evaluation results in substantial cost (experimental or computational). BO has achieved notable success in a wide range of applications, e.g., \citep{tang2024beacon, kudva2025multi, sorourifar2025adaptive}; however, BO often struggles in high-dimensional settings due to the curse of dimensionality \citep{binois2022survey}.
To mitigate this issue, \citep{eriksson2019scalable} proposed Trust-Region Bayesian Optimization (TuRBO), which proposes new evaluations by optimizing the acquisition function inside adaptive trust regions (TRs) while fitting GP hyperparameters on the accumulated data. Despite its empirical success, we identify a key limitation: \emph{trust-region restriction alone does not guarantee a well-calibrated GP complexity assumption.}

Recent work \citep{hvarfner2024vanilla} shows that overly complex GP priors harm high-dimensional BO and proposes scaling the (global) GP lengthscale with $\sqrt{D}$ to preserve kernel geometry. Inspired by this insight, we hypothesize that an analogous degeneracy arises in TuRBO’s \emph{local} GP models when lengthscales are not properly scaled with both dimension $D$ and trust region size $L$.
We validate this hypothesis using Maximum Information Gain (MIG) analysis and show that TuRBO’s local GP can still collapse toward near-independence in high dimensions. Motivated by this diagnosis, we derive a principled dimension- and TR-aware lengthscale scaling rule and propose a stabilized maximum a posteriori (MAP)-based hyperparameter estimation strategy.

Our main contributions in this work are as follows:
\begin{enumerate}
    \item We diagnose degeneracy of TuRBO’s local GP via MIG behavior across varying dimension $D$ and trust region size $L$.
    \item We derive a TR-aware lengthscale scaling rule and prove invariance of the MIG under relevant domain scaling.
    \item We propose \textbf{AdaScale-TuRBO}, which integrates a TR-aware LogNormal prior with MAP estimation to enforce proper scaling, achieving superior empirical performance on high-dimensional synthetic and real-world benchmarks.
\end{enumerate}

\section{BACKGROUND}

\subsection{Trust Region Bayesian Optimization}

BO is a sequential optimization strategy that relies on a probabilistic surrogate model (typically a GP) to learn a black-box function from observations. An acquisition function is used to select the next evaluation point by balancing exploration and exploitation; see Appendix~\ref{app:appendix_background_related} for additional details on GPs/BO as well as additional discussion on other related works. 

To address the curse of dimensionality in high-dimensional BO, \cite{eriksson2019scalable} proposed TuRBO, a local BO strategy. Instead of maintaining a single global surrogate model over the full search space scaled to $[0,1]^D$, TuRBO maintains one or more adaptive trust regions centered at incumbents and optimizes the acquisition function in each TR to sequentially propose new evaluation points. The TR size is adapted based on optimization progress, i.e., the TR is enlarged after $\tau_{\text{success}}$ consecutive successful rounds and shrunk after $\tau_{\text{fail}}$ consecutive failures. This mechanism encourages local refinement while still enabling exploration through TR adaptation.

\subsection{Maximum Information Gain}

The maximum information gain (MIG) \citep{srinivas2009gaussian} is widely used to quantify the \emph{assumed complexity} of a function under a GP prior.
Let $\bm X=\{\bm x_i\}_{i=1}^N \subset \mathcal X$ be a set of input points in the design space $\mathcal{X}$, let $\bm f_{\bm X}=(f(\bm x_i))_{i=1}^N$, and consider noisy observations $\bm y_{\bm X}=\bm f_{\bm X}+\bm\epsilon$ with $\bm\epsilon\sim\mathcal N(\bm 0,\sigma_\epsilon^2 \bm I)$.
Let $H(\cdot)$ denote (differential) Shannon entropy. The information gain (IG) is the mutual information between $\bm f_{\bm X}$ and $\bm y_{\bm X}$: 
\begin{align}
    \mathrm{IG}(\bm X) = H(\bm y_{\bm X}) - H(\bm y_{\bm X}\mid \bm f_{\bm X}).
\end{align}
Under a GP prior with (fixed) kernel hyperparameters and Gaussian noise variance $\sigma_\epsilon^2$, we have
\begin{align}
    \mathrm{IG}(\bm X)
    \;=\;
    \frac{1}{2}\log \left| \bm I + \sigma_\epsilon^{-2}\bm K_{\bm X\bm X} \right|,
\end{align}
where $\bm K_{\bm X\bm X}\in\mathbb R^{N\times N}$ is the kernel Gram matrix with entries $(\bm K_{\bm X\bm X})_{ij}=k(\bm x_i,\bm x_j)$.
The MIG over a domain $\mathcal X$ is then given by
\begin{align}
    \gamma_N(\mathcal X)
    \;=\;
    \max_{\bm X \subset \mathcal X:\, |\bm X|=N}
    \mathrm{IG}(\bm X).
\end{align}
Thus, $\gamma_N(\mathcal X)$ measures how much information the GP prior can extract from $N$ observations \emph{in the domain $\mathcal X$} (with the kernel and noise fixed).
In the near-independence regime $\bm K_{\bm X\bm X}\approx \bm I$, we obtain
\begin{align} \label{eq:mig-independent-simplified}
    \gamma_N(\mathcal X)\approx \frac{N}{2}\log(1+\sigma_\epsilon^{-2}),
\end{align}
which grows linearly with $N$. In this regime the prior assumes weak correlations between samples and is therefore effectively \emph{uninformative} about $f$ \citep{hvarfner2024vanilla}, making learning/optimization difficult.

\section{WHY TURBO CAN STILL FAIL IN HIGH DIMENSIONS}
\label{sec:MIG}

In this section, we demonstrate TuRBO’s local GP can still suffer from model degeneracy in high-dimensional settings due to improper lengthscale scaling.

\subsection{MIG behavior for local GP models.}

\cite{hvarfner2024vanilla} show that overly complex GP priors degrade the performance of vanilla BO in high dimensions. The core issue is geometric, i.e., in $[0,1]^D$, distances between randomly sampled points scale as $\Theta(\sqrt{D})$, causing kernel values to vanish unless the lengthscale scales accordingly. Without proper scaling, the kernel matrix approaches identity, leading to an effectively independent model.

TuRBO restricts modeling to a trust region $\mathcal{X}_L=[0,L]^D$, thereby reducing the search space. However, the expected distance between two uniformly sampled points in this domain still scales as $\Theta(L\sqrt{D})$, as shown in Proposition~\ref{proposition:expected-distance-bound} (proof in Appendix \ref{app:proofs}).

\begin{proposition} \label{proposition:expected-distance-bound}
Given a $D$-dimensional hypercube with side length $L$, the expected distance $\Delta(d)$ between two i.i.d.\ uniformly sampled points satisfies
\begin{align}
\frac{L}{3}\sqrt{D}
\leq
\Delta(d)
\leq
\frac{L}{\sqrt{6}}\sqrt{D}
\sqrt{\frac{1+2\sqrt{1-\frac{3}{5D}}}{3}}.
\end{align}
For $D \ge 20$, the upper bound is well approximated by $\frac{L}{\sqrt{6}}\sqrt{D}$, and thus both bounds scale as $\Theta(L\sqrt{D})$.
\end{proposition}

\begin{figure}[tb!]
  \centering
  \includegraphics[width=\linewidth]{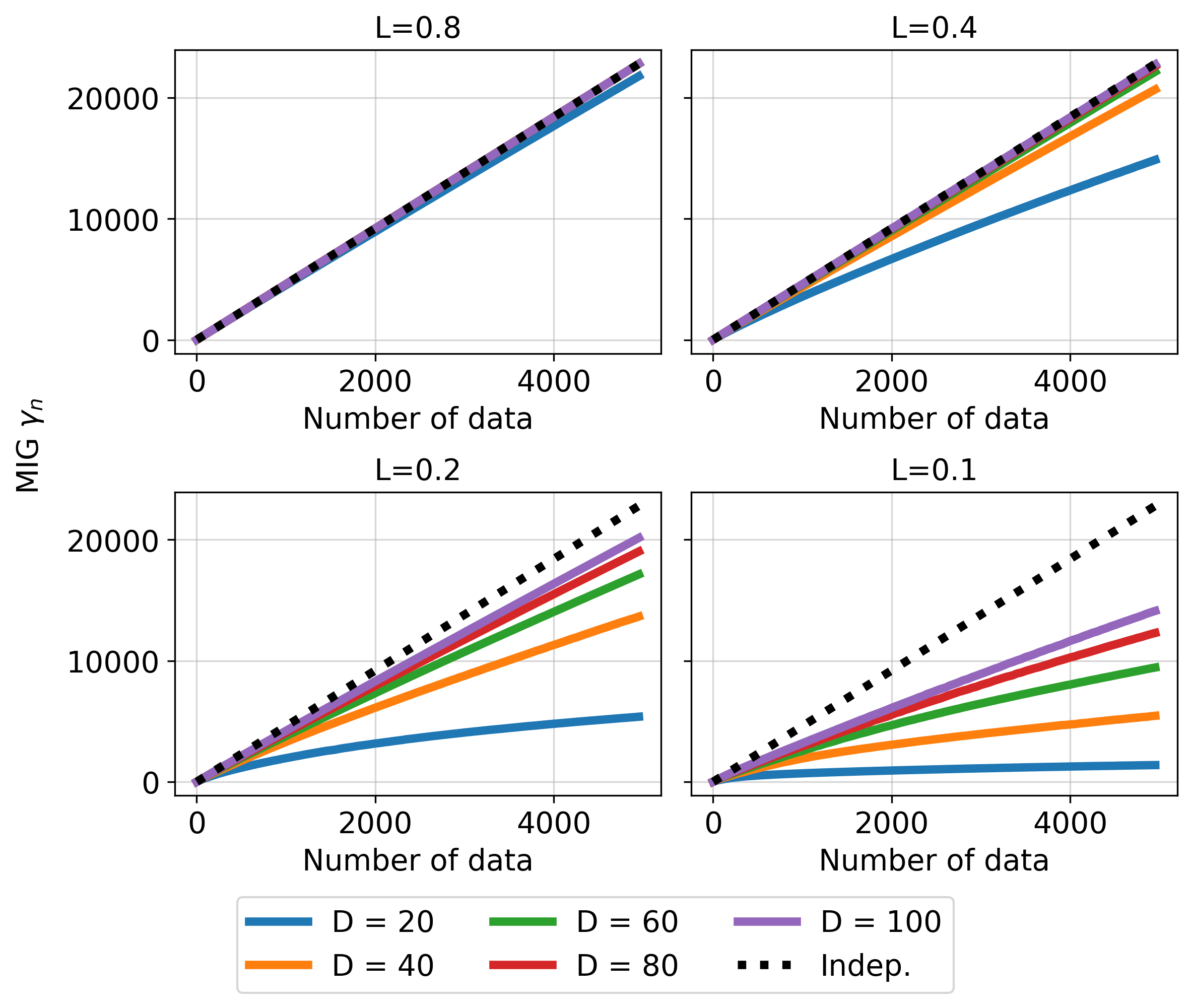}
  \caption{MIG scaling with number of observations. We consider TR side lengths $L=0.8,0.4,0.2,0.1$ and dimensions $D=20,40,60,80,100$ under an isotropic Mat\'ern-$5/2$ kernel with fixed lengthscale $\ell=0.5$. The dashed black line shows the independent-kernel regime ($\bm K=\bm I$). We estimate MIG using Sobol designs within the TR (a conservative proxy for the maximization).}
  \label{fig:MIG}
\end{figure}

This geometric scaling implies that the local GP in TuRBO may still collapse toward an independent model as $D$ increases unless the lengthscale is adjusted appropriately. To demonstrate this, we examine MIG as a function of the number of data points $N$, across varying $D$ and $L$ under a fixed isotropic lengthscale $\ell=0.5$. Figure~\ref{fig:MIG} shows, when $L=0.8$ or $0.4$, the MIG curve nearly overlaps with the linear dashed line (independent-kernel regime) for $D \geq 40$. This indicates that the local GP behaves almost independently and extracts very limited additional information from new samples.

When the TR shrinks ($L=0.2$ or $0.1$), MIG deviates from the linear regime. However, if the lengthscale is too large relative to the TR, the GP becomes overly smooth, resulting in an overly simplified complexity assumption for $f$. Such under-complex modeling may fail to capture non-smooth local structure, degrading optimization performance.

These observations reveal a critical issue: \emph{restricting the domain via a TR does not automatically ensure a properly calibrated complexity assumption.} Without principled scaling, the local GP may become either overly complex (near independence) or overly simple (excessively smooth), both of which destabilize acquisition optimization within the TR.

\subsection{Hyperparameter estimation instability.}

In practice, TuRBO estimates GP hyperparameters via maximum likelihood estimation (MLE) and specifies box constraints for the automatic relevance determination (ARD) lengthscales, e.g., $\ell \in [0.005,4]^D$ in the implementation in \texttt{BoTorch} \citep{balandat2020botorch}. In high dimensions, likelihood-based fitting can become brittle due to overparameterization (many lengthscales relative to available data), often producing extreme lengthscales and poorly calibrated uncertainty \citep{hvarfner2024vanilla}. This motivates replacing pure MLE with a more stable MAP approach that incorporates a dimension- and TR-aware scaling prior for TuRBO's local GP model.

\section{PROPOSED METHOD} 
\label{sec:Method}

Section~\ref{sec:MIG} shows that TuRBO’s local GP can be overly complex or overly simple in high dimensions under fixed lengthscale design. We now derive a principled scaling rule that preserves kernel geometry and prior complexity across both dimension $D$ and TR size $L$.

For $\mathcal{X}_L=[0,L]^D$, typical pairwise distances scale as $\Theta(L\sqrt{D})$ (cf.\ Proposition~\ref{proposition:expected-distance-bound}). Therefore, to prevent the kernel from degenerating toward independence within the TR, the lengthscale should scale proportionally to $L\sqrt{D}$, preserving kernel geometry as $D$ and $L$ change.

\begin{proposition} \label{proposition2}
Let $\mathcal X_L=[0,L]^D$ for any $L \in [0,1]$.
Consider an isotropic stationary kernel $k_\ell(\bm x,\bm x')=\kappa\!\left(\frac{\|\bm x-\bm x'\|}{\ell}\right)$.
If the global GP uses $\ell \propto \sqrt{D}$ and the local GP uses $\ell \propto L\sqrt{D}$, then for all $N\ge 1$,
\begin{align}
\gamma_N(\mathcal X_1, k_{\ell\propto\sqrt{D}})
=
\gamma_N(\mathcal X_L, k_{\ell\propto L\sqrt{D}}).
\end{align}
\end{proposition}

The proof is given in Appendix \ref{app:proofs}. This result shows that scaling $\ell \propto L\sqrt{D}$ ensures that the local GP maintains the same effective complexity assumption as a properly scaled global GP \citep{hvarfner2024vanilla}.

To enforce this scaling in practice, we propose the following LogNormal prior for the local GP lengthscales:
\begin{align}\label{eq:lognormal-prior}
\ell_i \sim
\mathcal{LN}\!\left(
\mu_0 + \log(L\sqrt{D}),
\sigma_0
\right),
\end{align}
where $(\mu_0,\sigma_0)$ are the fixed base parameters reported in \citep{hvarfner2024vanilla}, i.e., $\mu_0 = \sqrt{2}$ and $\sigma_0 = \sqrt{3}$. This prior shifts the mean and mode of the lengthscale distribution by a multiplicative factor of $L\sqrt{D}$, ensuring dimension- and TR-consistent scaling.

Following \citep{hvarfner2024vanilla}, we fix the signal variance to $\sigma_f^2=1$ to avoid variance-lengthscale coupling during hyperparameter estimation. Instead of pure MLE, we suggest the use of MAP estimation to stabilize training in high-dimensional settings.

In summary, we modify TuRBO by replacing MLE with MAP and embedding the proposed prior in \eqref{eq:lognormal-prior} when fitting the GP hyperparameters in each iteration. We refer to the resulting algorithm as \textbf{AdaScale-TuRBO}. Complete pseudocode for AdaScale-TuRBO is provided in Appendix \ref{app:pseudocode-adascaleturbo}.

\section{EXPERIMENTS} 
\label{sec:experiment}

In this section, we compare AdaScale-TuRBO against several high-dimensional BO baselines: vanilla BO with a dimension-scaled lengthscale prior (D-scaled LogEI) \citep{hvarfner2024vanilla}, TuRBO \citep{eriksson2019scalable}, and a recent Linear BO \citep{doumont2025we}. We also include an additional baseline that combines TuRBO with the dimension-scaled lengthscale prior of \citep{hvarfner2024vanilla}, which we refer to as D-scaled TuRBO. In this work, we use an ARD Mat\'ern-$5/2$ kernel for all methods except Linear BO, which uses a linear kernel by default. To ensure a fair comparison, all methods employ the LogEI acquisition function \citep{ament2023unexpected}, optimized using L-BFGS-B with five multi-start restarts. Each method is initialized with the same 10 Sobol samples. The code can be found at this GitHub link: \url{https://github.com/PaulsonLab/AdaScale-TuRBO.git}.

\subsection{Synthetic Benchmark}

We consider three standard synthetic benchmark functions widely used in the BO literature: Schwefel, Rastrigin, and Michalewicz. The mathematical expressions are provided in Appendix \ref{app:experiment-details}. We evaluate performance on 50-dimensional and 100-dimensional instances, with total budgets of 500 and 1{,}000 function evaluations, respectively. To reduce computational overhead from GP fitting ($\mathcal{O}(N^3)$ complexity), we refit the GP every 10 iterations for all methods except Linear BO, whose inference cost only scales as $\mathcal{O}(D)$ and therefore can easily be refit every iteration. 

\begin{figure}[tb!]
  \centering
  \includegraphics[width=\linewidth]{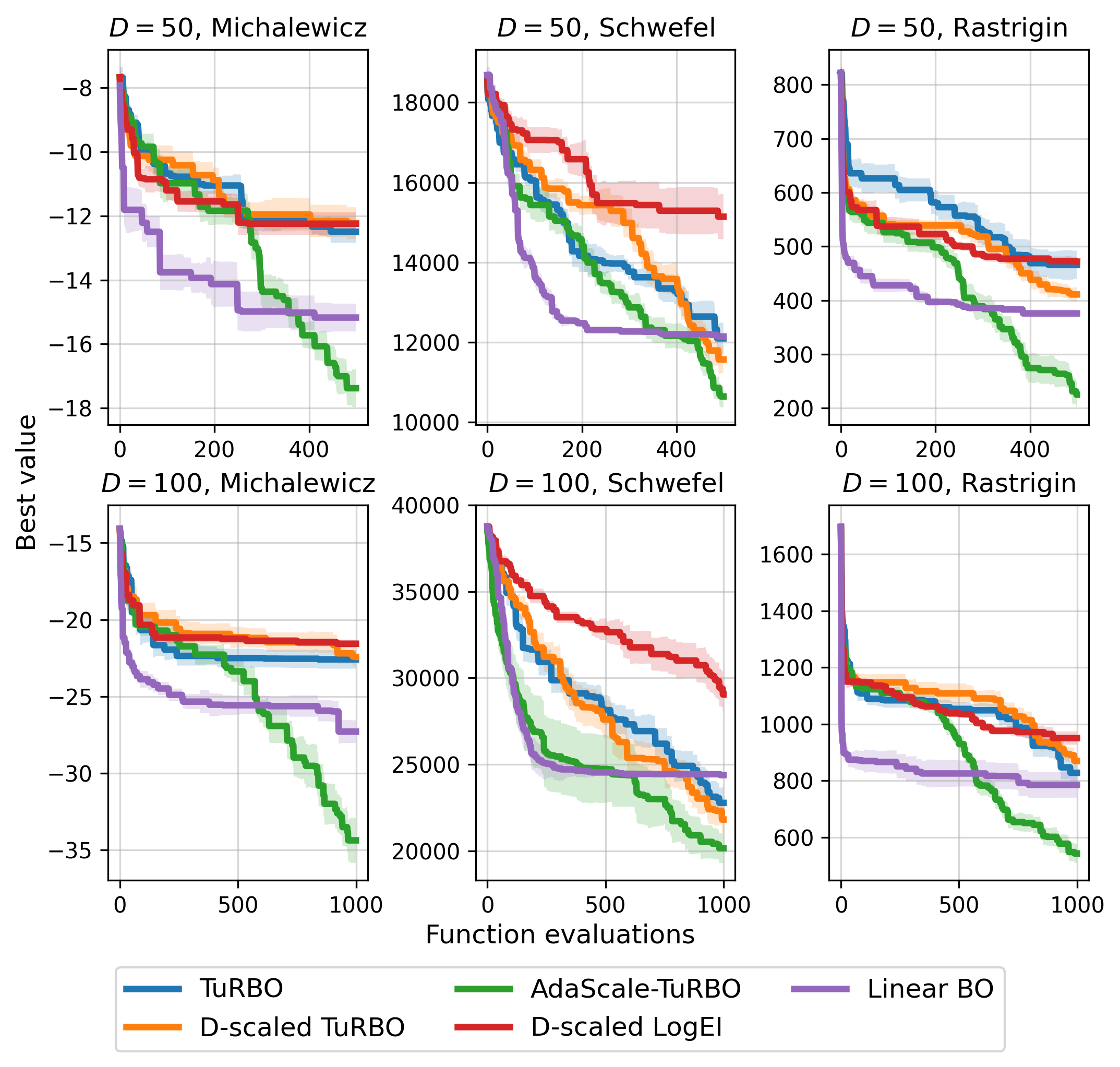}
  \caption{Best observed function value for synthetic functions. The first and second rows correspond to $50$-dimensional and $100$-dimensional problems. The median and standard error are reported across 10 replicates for all methods.}
  \label{fig:Synthetic}
\end{figure}

Figure~\ref{fig:Synthetic} reports the best observed function value over iterations. AdaScale-TuRBO consistently outperforms all baselines across all functions and dimensions. In particular, it substantially improves upon TuRBO, demonstrating that calibrating the local GP complexity via \eqref{eq:lognormal-prior} can significantly enhance optimization performance. Interestingly, D-scaled TuRBO performs similarly to vanilla TuRBO, suggesting that scaling only with $\sqrt{D}$ -- without incorporating TR size $L$ -- can make the local GP overly smooth within small TRs and degrade performance, consistent with the analysis in Section~\ref{sec:MIG}.
Additional synthetic experiments and the violin plots of the final best-found values across 10 replicates are shown in Appendix \ref{app:add-experiments-results}. 

\subsection{Real-World Benchmark}

\begin{figure}[tb!]
  \centering
  \includegraphics[width=\linewidth]{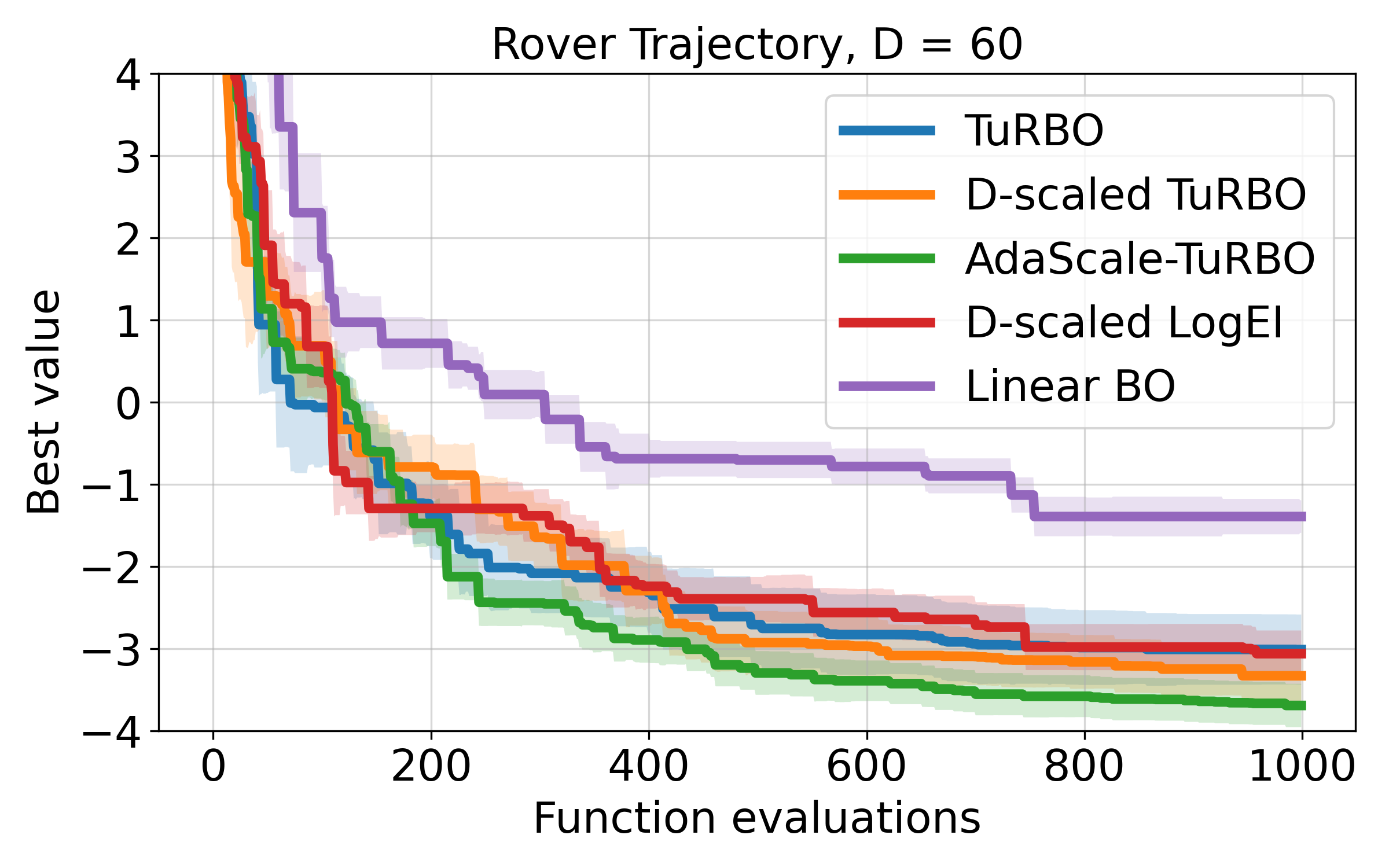}
  \caption{Best observed function value for the 60-dimensional rover trajectory planning problem. The median and standard error are reported across 10 replicates for all methods.}
  \label{fig:Rover}
\end{figure}

We further evaluate performance on a 60-dimensional rover trajectory planning problem \citep{wang2018batched}, a widely studied benchmark in high-dimensional BO. The experimental protocol matches that of the synthetic experiments. 
Figure~\ref{fig:Rover} shows that AdaScale-TuRBO achieves the best performance among all methods. In particular, it significantly improves upon TuRBO, confirming that controlling local GP complexity through the proposed dimension- and TR-aware lengthscale prior \eqref{eq:lognormal-prior} yields substantial gains on a real-world high-dimensional optimization task.

\section{CONCLUSIONS}
\label{sec:conclusions}

In this work, we show that local GP model degeneracy is a key factor that limits the performance of TuRBO in high-dimensional optimization (especially in problems with $> 50$ dimensions). By analyzing the maximum information gain (MIG), we 
demonstrate that fixed lengthscales can cause TuRBO’s local GP to become either overly complex (near independence) or overly smooth as the dimension and trust region size vary. We then derive a principled scaling rule showing that the lengthscale should scale with $L\sqrt{D}$ to preserve kernel geometry and maintain consistent model complexity. Building on this insight, AdaScale-TuRBO achieves significantly improved performance over TuRBO and other popular high-dimensional BO methods on both synthetic and real-world benchmarks.


\bibliographystyle{abbrvnat}
\bibliography{reference}

\section*{Checklist}



\begin{enumerate}

  \item For all models and algorithms presented, check if you include:
  \begin{enumerate}
    \item A clear description of the mathematical setting, assumptions, algorithm, and/or model. [Yes]
    \item An analysis of the properties and complexity (time, space, sample size) of any algorithm. [Yes]
    \item (Optional) Anonymized source code, with specification of all dependencies, including external libraries. [Yes]
  \end{enumerate}

  \item For any theoretical claim, check if you include:
  \begin{enumerate}
    \item Statements of the full set of assumptions of all theoretical results. [Yes]
    \item Complete proofs of all theoretical results. [Yes]
    \item Clear explanations of any assumptions. [Yes]     
  \end{enumerate}

  \item For all figures and tables that present empirical results, check if you include:
  \begin{enumerate}
    \item The code, data, and instructions needed to reproduce the main experimental results (either in the supplemental material or as a URL). [Yes]
    \item All the training details (e.g., data splits, hyperparameters, how they were chosen). [Yes]
    \item A clear definition of the specific measure or statistics and error bars (e.g., with respect to the random seed after running experiments multiple times). [Yes]
    \item A description of the computing infrastructure used. (e.g., type of GPUs, internal cluster, or cloud provider). [Yes]
  \end{enumerate}

  \item If you are using existing assets (e.g., code, data, models) or curating/releasing new assets, check if you include:
  \begin{enumerate}
    \item Citations of the creator If your work uses existing assets. [Yes]
    \item The license information of the assets, if applicable. [Not Applicable]
    \item New assets either in the supplemental material or as a URL, if applicable. [Yes]
    \item Information about consent from data providers/curators. [Not Applicable]
    \item Discussion of sensible content if applicable, e.g., personally identifiable information or offensive content. [Not Applicable]
  \end{enumerate}

  \item If you used crowdsourcing or conducted research with human subjects, check if you include:
  \begin{enumerate}
    \item The full text of instructions given to participants and screenshots. [Not Applicable]
    \item Descriptions of potential participant risks, with links to Institutional Review Board (IRB) approvals if applicable. [Not Applicable]
    \item The estimated hourly wage paid to participants and the total amount spent on participant compensation. [Not Applicable]
  \end{enumerate}

\end{enumerate}

\clearpage
\appendix
\numberwithin{equation}{section}
\numberwithin{figure}{section}
\numberwithin{table}{section}

\thispagestyle{empty}

\onecolumn
\aistatstitle{Supplementary Material}

\section{Background and Related Work}
\label{app:appendix_background_related}

\subsection{Bayesian Optimization and Gaussian Processes}
\label{sec:bo_gp}

Bayesian optimization (BO) is a sequential strategy for minimizing an expensive black-box objective
$f:\mathcal X \rightarrow \mathbb R$. In this paper we take $\mathcal X_1=[0,1]^D$ as the global domain and use $\mathcal X_L=[0,L]^D$ to denote a trust region (TR) hypercube of side length $L$. At iteration $t$, BO queries $\bm x_t\in\mathcal X$ and observes
\begin{align}
    y_t = f(\bm x_t) + \epsilon_t,\qquad \epsilon_t \sim \mathcal N(0,\sigma_\epsilon^2),
\end{align}
forming the dataset $\mathcal D_t=\{(\bm x_i,y_i)\}_{i=1}^t$.

A common surrogate model in BO is a Gaussian process (GP) prior
$f \sim \mathcal{GP}(m(\cdot),k(\cdot,\cdot))$, where $m(\cdot)$ is the mean function and $k(\cdot,\cdot)$ is the covariance or kernel function (that must be positive definite). Given inputs $\bm X_t=[\bm x_1,\ldots,\bm x_t]^\top$ and observations $\bm y_t=[y_1,\ldots,y_t]^\top$, define the Gram matrix $\bm K_t=\bm K_{\bm X_t\bm X_t}$ with entries $(\bm K_t)_{ij}=k(\bm x_i,\bm x_j)$ and the cross-covariance vector $\bm k_t(\bm x)=[k(\bm x_1,\bm x),\ldots,k(\bm x_t,\bm x)]^\top$. The GP posterior mean and variance are
\begin{align}
    \mu_t(\bm x) &= \bm k_t(\bm x)^\top (\bm K_t+\sigma_\epsilon^2\bm I)^{-1}\bm y_t,\\
    \sigma_t^2(\bm x) &= k(\bm x,\bm x) - \bm k_t(\bm x)^\top(\bm K_t+\sigma_\epsilon^2\bm I)^{-1}\bm k_t(\bm x).
\end{align}

In our experiments we use the automatic relevance determination (ARD) Mat\'ern-$5/2$ kernel, i.e., 
\begin{align}
k(\bm x,\bm x')
= \sigma_f^2\Big(1+\sqrt{5}\,r+\tfrac{5}{3}r^2\Big)\exp(-\sqrt{5}\,r),
\qquad
r=\sqrt{\sum_{j=1}^D \frac{(x_j-x'_j)^2}{\ell_j^2}},
\end{align}
with hyperparameters given by the lengthscales $\bm\ell=(\ell_1,\ldots,\ell_D)$, signal variance $\sigma_f^2$, and noise variance $\sigma_\epsilon^2$.

At each iteration, BO selects the next query point by optimizing an acquisition function $\alpha_t$ built from $(\mu_t,\sigma_t)$:
\begin{align}
    \bm x_{t+1} \in \argmax_{\bm x \in \mathcal X} \alpha_t(\bm x).
\end{align}
Common choices include expected improvement (EI) \citep{jones1998efficient} (and its numerically stabilized variant LogEI \citep{ament2023unexpected}) and upper confidence bound (UCB) \citep{srinivas2009gaussian}. 
Hyperparameters are typically fit by maximizing the GP marginal log likelihood (MLE) or by adding a prior and performing MAP estimation, which can be particularly helpful in high-dimensional settings where training can be sensitive to initialization and priors \citep{hvarfner2024vanilla,xu2024standard,papenmeier2025understanding}.

\subsection{Related Work on High-Dimensional Bayesian Optimization}
\label{sec:related_work}

High-dimensional BO (HDBO) has been studied extensively in recent years, motivated by the empirical difficulty of fitting and optimizing GP surrogates as $D$ grows \citep{binois2022survey}. Existing methods can be roughly grouped into three directions.

\paragraph{Structure-exploiting methods (effective low dimension).}
A large body of HDBO work assumes the objective has low-dimensional structure. Examples include additive decompositions \citep{kandasamy2015high,gardner2017discovering,ziomek2023random} and embedded or sparse axis-aligned subspaces \citep{nayebi2019framework,eriksson2021high,song2022monte,hellsten2023high}. When these assumptions match the objective, such methods can significantly reduce the effective search space and improve sample efficiency. However, performance can degrade when the assumed structure is violated or the embedding misses key directions \citep{nayebi2019framework}.

\paragraph{``Vanilla'' BO with dimension-aware priors and initialization.}
Recent work argues that standard GP-based BO can remain competitive in high dimensions without explicit structural assumptions, provided that the GP prior and training are carefully calibrated. \cite{hvarfner2024vanilla} interpret HDBO failures through prior complexity and propose scaling the GP lengthscale prior with $\sqrt{D}$ to preserve kernel geometry. \cite{xu2024standard} identify a related failure mode during GP training and propose simple initialization strategies that mitigate vanishing gradients, while \cite{papenmeier2025understanding} analyze why recent recipes work and emphasize the interaction between initialization, local search behavior, and acquisition optimization. Complementary work shows that simple linear models with appropriate geometric transformations can be surprisingly strong on HDBO benchmarks \citep{doumont2025we}.

\paragraph{Local Bayesian optimization.}
Another prominent direction focuses on local modeling and search to reduce global high-dimensional challenges. Trust region-based BO, beginning with TuRBO \citep{eriksson2019scalable}, restricts acquisition optimization to adaptive local regions and has inspired many variants \citep{chen2025enhancing,namura2025regional,eriksson2021scalable,tang2024tr,daulton2022multi}. Beyond TR-based methods, other local BO approaches leverage local first-order information inferred from the GP posterior \citep{muller2021local,tang2024cages,fan2024minimizing,stenger2025local,nguyen2022local}, or incorporate second-order information for faster local refinement \citep{tang2025nest,brunzema2026bayesqp}. Our work fits within this local-BO line: we focus on calibrating the \emph{local} GP prior complexity inside a TR by scaling lengthscales with both $D$ and the TR size $L$.

\section{Proofs}
\label{app:proofs}

\subsection{Independent kernel information gain}
\label{sec:proof_indep_mig}

In the ``independent kernel'' regime, the Gram matrix satisfies
$\bm K_{\bm X\bm X} \approx \bm I_N$.
When $\bm K_{\bm X\bm X} = \bm I_N$ exactly, the information gain reduces to:
\begin{align}
\mathrm{IG}(\bm y_{\bm X}; f)
&= \frac{1}{2}\log\!\left|\bm I_N + \sigma_\epsilon^{-2}\bm K_{\bm X\bm X}\right|
 = \frac{1}{2}\log\!\left|\bm I_N + \sigma_\epsilon^{-2}\bm I_N\right| \nonumber\\
&= \frac{1}{2}\log\!\left|(1+\sigma_\epsilon^{-2})\bm I_N\right|
 = \frac{1}{2}\log\!\left((1+\sigma_\epsilon^{-2})^N\right)
 = \frac{N}{2}\log(1+\sigma_\epsilon^{-2}).
\end{align}
Since this value is independent of the particular design set $\bm X$ when
$\bm K_{\bm X\bm X}=\bm I_N$, the corresponding MIG equals
$\gamma_N = \frac{N}{2}\log(1+\sigma_\epsilon^{-2})$. This proves the statement shown in \eqref{eq:mig-independent-simplified}. 

\subsection{Proof of Proposition~\ref{proposition:expected-distance-bound}}
\label{sec:proof_prop1}

Let $\mathcal{X}_1=[0,1]^D$ and let $\bm x,\bm x' \overset{\mathrm{i.i.d.}}{\sim} \mathrm{Unif}(\mathcal X_1)$.
Define the mean distance on the unit hypercube as
$M_D := \mathbb E\|\bm x-\bm x'\|_2$.
In \citep{anderssen1976concerning}, they establish the following bounds:
\begin{align} \label{eq:unit_cube_bound}
\frac{1}{3}\sqrt{D}
\;\le\;
M_D
\;\le\;
\sqrt{\frac{D}{6}}
\sqrt{\frac{1+2\sqrt{1-\frac{3}{5D}}}{3}}.
\end{align}

Now consider $\mathcal X_L=[0,L]^D$ 
Define the bijection $\phi:\mathcal X_1\to\mathcal X_L$ by $\phi(\bm x)=L\bm x$.
Then if $\bm x,\bm x'\sim\mathrm{Unif}(\mathcal X_1)$, the transformed variables
$\bm u=\phi(\bm x)$ and $\bm u'=\phi(\bm x')$ are i.i.d.\ $\mathrm{Unif}(\mathcal X_L)$.
By homogeneity of the Euclidean norm, we also have:
\begin{align}
\|\bm u-\bm u'\|_2
= \|L\bm x-L\bm x'\|_2
= L\|\bm x-\bm x'\|_2.
\end{align}
Taking expectations yields
\begin{align}
\Delta(d) := \mathbb E\|\bm u-\bm u'\|_2
= L\,\mathbb E\|\bm x-\bm x'\|_2
= L M_D.
\end{align}
Multiplying both sides of \eqref{eq:unit_cube_bound} by $L$ gives
\begin{align}
\frac{L}{3}\sqrt{D}
\;\le\;
\Delta(d)
\;\le\;
\frac{L}{\sqrt{6}}\sqrt{D}
\sqrt{\frac{1+2\sqrt{1-\frac{3}{5D}}}{3}},
\end{align}
which is exactly the bounds stated in Proposition~\ref{proposition:expected-distance-bound}. Finally, since $\sqrt{\frac{1+2\sqrt{1-\frac{3}{5D}}}{3}} \to 1$ as $D \to \infty$, we also see that $\Delta(d) = \Theta(L\sqrt{D})$ in the large limit of $D$, which completes the proof. $\hfill \square$

\subsection{Proof of Proposition~\ref{proposition2}}
\label{sec:proof_prop2}

Fix $D\ge 1$ and $L\in(0,1]$. Let $\mathcal X_1=[0,1]^D$ and $\mathcal X_L=[0,L]^D$.
Consider an isotropic stationary kernel of the form
$k_\ell(\bm x,\bm x')=\kappa\!\left(\frac{\|\bm x-\bm x'\|_2}{\ell}\right)$.
Let the global and local lengthscales share the same constant of proportionality $c>0$:
\begin{align*}
\ell_1 = c\sqrt{D},
\qquad
\ell_L = cL\sqrt{D}.    
\end{align*}
Assume both models use the same Gaussian observation noise variance $\sigma_\epsilon^2$.

Define the bijection $\phi:\mathcal X_1\to\mathcal X_L$ by $\phi(\bm x)=L\bm x$.
Fix any $N$-point design $\bm X=\{\bm x_1,\ldots,\bm x_N\}\subset \mathcal X_1$ and let
$\bm U=\phi(\bm X)=\{\bm u_1,\ldots,\bm u_N\}\subset \mathcal X_L$ with $\bm u_i=L\bm x_i$.
Let $\bm f_{\bm X}=(f(\bm x_i))_{i=1}^N$ under the GP prior $\mathcal{GP}(0,k_{\ell_1})$ on $\mathcal X_1$,
and let $\bm f_{\bm U}=(f(\bm u_i))_{i=1}^N$ under the GP prior $\mathcal{GP}(0,k_{\ell_L})$ on $\mathcal X_L$.
Then for any $i,j$,
\begin{align}
k_{\ell_L}(\bm u_i,\bm u_j)
&=\kappa\!\left(\frac{\|L\bm x_i-L\bm x_j\|_2}{cL\sqrt{D}}\right)
=\kappa\!\left(\frac{\|\bm x_i-\bm x_j\|_2}{c\sqrt{D}}\right)
= k_{\ell_1}(\bm x_i,\bm x_j).
\end{align}
Hence, the Gram matrices coincide:
\[
\bm K^{(\ell_L)}_{\bm U\bm U}=\bm K^{(\ell_1)}_{\bm X\bm X}.
\]
Thus, the finite-dimensional marginals satisfy
$\bm f_{\bm U}\sim \mathcal N(\bm 0,\bm K^{(\ell_L)}_{\bm U\bm U})
= \mathcal N(\bm 0,\bm K^{(\ell_1)}_{\bm X\bm X})\sim \bm f_{\bm X}$.
With identical noise variance, the corresponding observation vectors
$\bm y_{\bm U}=\bm f_{\bm U}+\bm\epsilon$ and $\bm y_{\bm X}=\bm f_{\bm X}+\bm\epsilon$
also have the same joint Gaussian distribution with their respective latent vectors.

Since the information gain is
$\mathrm{IG}(\bm X) := I(\bm y_{\bm X};\bm f_{\bm X})$ (where $I(\cdot)$ denotes the mutual information) and likewise for $\bm U$, we conclude
\begin{align}
I(\bm y_{\bm U};\bm f_{\bm U})
= \frac{1}{2}\log\!\left|\bm I_N+\sigma_\epsilon^{-2}\bm K^{(\ell_L)}_{\bm U\bm U}\right|
= \frac{1}{2}\log\!\left|\bm I_N+\sigma_\epsilon^{-2}\bm K^{(\ell_1)}_{\bm X\bm X}\right|
= I(\bm y_{\bm X};\bm f_{\bm X}).
\end{align}

Finally, because $\phi$ is a bijection, it induces a one-to-one correspondence between
$N$-point subsets of $\mathcal X_1$ and $N$-point subsets of $\mathcal X_L$.
Maximizing over designs yields
\[
\gamma_N(\mathcal X_1,k_{\ell_1})=\gamma_N(\mathcal X_L,k_{\ell_L}),
\qquad \forall N\ge 1,
\]
which completes the proof. $\hfill \square$

\section{Pseudocode for AdaScale-TuRBO}
\label{app:pseudocode-adascaleturbo}

Algorithm~\ref{alg:adascale-turbo} follows the TuRBO-1 template \citep{eriksson2019scalable}: it maintains a trust region (TR) centered at the current incumbent, proposes new evaluations by optimizing an acquisition function restricted to the TR, and adaptively expands/shrinks the TR based on consecutive success/failure counts. When the TR side length falls below a minimum threshold, the local run is restarted.

\paragraph{What changes relative to TuRBO.}
AdaScale-TuRBO differs from the original TuRBO implementation in two places.
First, we fit the local GP hyperparameters using MAP with the TR-aware LogNormal lengthscale prior in \eqref{eq:lognormal-prior} (and fix $\sigma_f^2=1$ as described in the main text), rather than pure MLE with box constraints. Second, for fair comparison across baselines, we use $\mathrm{LogEI}$ \citep{ament2023unexpected} as the default acquisition function in all experiments; any standard acquisition can be used in its place. The original TuRBO commonly uses Thompson sampling for candidate selection, especially in batch settings.

\paragraph{Default TuRBO hyperparameters.}
Unless otherwise stated, we use the TuRBO defaults \citep{eriksson2019scalable}: success tolerance $\tau_{\text{succ}}=3$,  failure tolerance $\tau_{\text{fail}}=\left\lceil \max\!\left(\frac{4}{q},\frac{D}{q}\right)\right\rceil$ (batch size $q$), initial TR side length $L_0=0.8$, minimum $L_{\min}=0.5^7$, and maximum $L_{\max}=1.6$. We initialize each restart with $N_0$ space-filling points with $N_0 = 10$.  

\paragraph{Ablations.}
We provide an ablation on the initial TR size $L_0$ in Appendix \ref{sec:ablation_L0}, and show that AdaScale-TuRBO consistently improves over TuRBO across a range of $L_0$ choices.

\begin{algorithm}[t]
\caption{AdaScale-TuRBO}\label{alg:adascale-turbo}
\begin{algorithmic}[1]
\Require Dimension $D$; domain $\mathcal{X}=[0,1]^D$; black-box objective $f$ (minimize);
initial design size $N_0$; total evaluation budget $T$; batch size $q$;
TR hyperparameters $L_0, L_{\min}, L_{\max}, \tau_{\text{succ}}, \tau_{\text{fail}}$.
\Ensure Best-found point $\bm x^\star$.
\State Initialize a space-filling design $\{\bm x_i\}_{i=1}^{N_0} \subset \mathcal{X}$ and evaluate $y_i \gets f(\bm x_i)$.
\State $\mathcal{D} \gets \{(\bm x_i,y_i)\}_{i=1}^{N_0}$; \quad $n \gets N_0$ \Comment{$n$ counts total evaluations}
\State $L \gets L_0$; \quad $\textsf{succ} \gets 0$; \quad $\textsf{fail} \gets 0$
\State \textbf{Fit} local GP on $\mathcal{D}$ via MAP using the prior in Eq.~\eqref{eq:lognormal-prior}
\While{$n < T$}
    \State $(\bm x^\star, y^\star) \gets \argmin_{(\bm x,y)\in \mathcal{D}} y$ \Comment{Incumbent}
    \State $\mathcal{X}_{\mathrm{TR}} \gets \big([\bm x^\star-\tfrac{L}{2},\,\bm x^\star+\tfrac{L}{2}]\big)\cap \mathcal{X}$
    \State Choose acquisition $\alpha(\cdot)$ (default: $\mathrm{LogEI}$) and propose
    $\{\bm x^{(j)}\}_{j=1}^{q} \gets \argmax_{\{ \bm{x}^{(j)} \}_{j=1}^q \in \mathcal{X}_{\mathrm{TR}}^q} \alpha(\{ \bm{x}^{(j)} \}_{j=1}^q ;\mathcal{D})$
    \State Evaluate $y^{(j)} \gets f(\bm x^{(j)})$ for $j=1,\dots,q$ and set $\mathcal{D} \gets \mathcal{D}\cup\{(\bm x^{(j)},y^{(j)})\}_{j=1}^q$
    \State $n \gets n + q$
    \State \textbf{Refit/update} GP via MAP on $\mathcal{D}$ using Eq.~\eqref{eq:lognormal-prior}

    \State $y_{\min} \gets \min_{j=1,\dots,q} y^{(j)}$
    \If{$y_{\min} < y^\star$} \Comment{Success (optionally with a small relative threshold as in TuRBO)}
        \State $\textsf{succ} \gets \textsf{succ}+1$; \quad $\textsf{fail} \gets 0$
    \Else
        \State $\textsf{succ} \gets 0$; \quad $\textsf{fail} \gets \textsf{fail}+1$
    \EndIf

    \If{$\textsf{succ} \ge \tau_{\text{succ}}$}
        \State $L \gets \min(L_{\max},\, 2L)$; \quad $\textsf{succ} \gets 0$
    \ElsIf{$\textsf{fail} \ge \tau_{\text{fail}}$}
        \State $L \gets \tfrac{1}{2}L$; \quad $\textsf{fail} \gets 0$
    \EndIf

    \If{$L < L_{\min}$}
        \State \textbf{Restart:} $L \gets L_0$; \ $\textsf{succ}\gets 0$; \ $\textsf{fail}\gets 0$
        \State Draw a new space-filling design $\{\bm x_i\}_{i=1}^{N_0}\subset \mathcal{X}$, evaluate $y_i\gets f(\bm x_i)$
        \State $\mathcal{D} \gets \{(\bm x_i,y_i)\}_{i=1}^{N_0}$; \quad $n \gets n + N_0$
        \State \textbf{Fit} local GP on $\mathcal{D}$ via MAP using Eq.~\eqref{eq:lognormal-prior}
    \EndIf
\EndWhile
\State \Return $\bm x^\star$
\end{algorithmic}
\end{algorithm}

\section{Experiment Details}
\label{app:experiment-details}

\subsection{Implementation}
\label{sec:implementation}

We compare AdaScale-TuRBO to four baselines: TuRBO, D-scaled TuRBO, D-scaled LogEI, and Linear BO. All GP-based methods are implemented in \texttt{BoTorch} \citep{balandat2020botorch} and \texttt{GPyTorch} \citep{gardner2018gpytorch} and use a Mat\'ern-$5/2$ kernel unless otherwise noted. The code can be found on GitHub: \url{https://github.com/PaulsonLab/AdaScale-TuRBO.git}. 

\begin{itemize}
    \item \textit{TuRBO:} We implement TuRBO-1 using the official BoTorch tutorial implementation.\footnote{\url{https://botorch.org/docs/tutorials/turbo_1/}}
    \item \textit{D-scaled TuRBO:} Starting from the TuRBO implementation above, we replace the covariance module with a dimension-scaled LogNormal lengthscale prior as in \citet{hvarfner2024vanilla}. Concretely, we swap the default \texttt{covar\_module} with \texttt{get\_covar\_module\_with\_dim\_scaled\_prior}.
    \item \textit{AdaScale-TuRBO:} We implement AdaScale-TuRBO by modifying the same TuRBO code path, but using the proposed TR-aware LogNormal prior in \eqref{eq:lognormal-prior} (i.e., shifting the prior by $\log(L\sqrt{D})$). We fit GP hyperparameters using MAP rather than pure MLE.
    \item \textit{D-scaled LogEI:} We implement vanilla BO with the dimension-scaled lengthscale prior following the BoTorch examples,\footnote{\url{https://botorch.org/}} using a Mat\'ern-$5/2$ kernel and the LogEI acquisition function.
    \item \textit{Linear BO:} We use the authors' public implementation.\footnote{\url{https://github.com/colmont/linear-bo}}
\end{itemize}

For all GP-based methods, we train using BoTorch's \texttt{fit\_gpytorch\_mll} routine. Acquisition functions are optimized with \texttt{optimize\_acqf} using \texttt{num\_restarts}=5 and \texttt{raw\_samples}=20 (L-BFGS-B). Each method is initialized with 10 Sobol points. To reduce GP training overhead, we refit GP hyperparameters every 10 iterations for all GP-based methods except Linear BO (which is refit every iteration, as described in the main text).

\subsection{Computing Resources}
\label{sec:compute}

All experiments were run on the Ohio Supercomputer Center (OSC) \textit{Cardinal} cluster. We used CPU-only dense compute nodes with 2$\times$ Intel Xeon CPU Max 9470 processors and 512 GB DDR5 RAM (plus 128 GB HBM2e) per node.

\subsection{Synthetic Functions}
\label{sec:synthetic_expression}

This subsection provides the test functions used in Section~\ref{sec:experiment}. In all synthetic experiments, BO operates over the normalized domain $\mathcal X_1=[0,1]^D$. For a function with native bounds $\bm a \le \bm x \le \bm b$ (componentwise), we map $\bm z\in[0,1]^D$ to the native domain via the affine transform
\begin{align}
\bm x(\bm z) = \bm a + (\bm b-\bm a)\odot \bm z,
\end{align}
and evaluate $f(\bm x(\bm z))$, where $\odot$ denotes the elementwise product. 

\paragraph{Schwefel.}
The Schwefel function is
\begin{align}
f(\bm x) = 418.9829\,D - \sum_{i=1}^D x_i \sin\!\big(\sqrt{|x_i|}\big),
\qquad x_i \in [-500,500].
\end{align}
It has global minimum $f(\bm x^\star)=0$ at $x_i^\star \approx 420.9687$ for all $i$.\footnote{\url{https://www.sfu.ca/~ssurjano/schwef.html}}

\paragraph{Rastrigin.}
The Rastrigin function is
\begin{align}
f(\bm x) = 10\,D + \sum_{i=1}^D \Big(x_i^2 - 10\cos(2\pi x_i)\Big),
\qquad x_i \in [-5.12,5.12],
\end{align}
with global minimum $f(\bm x^\star)=0$ at $\bm x^\star=\bm 0$.\footnote{\url{https://www.sfu.ca/~ssurjano/rastr.html}}

\paragraph{Michalewicz.}
The Michalewicz function is
\begin{align}
f(\bm x) = -\sum_{i=1}^D \sin(x_i)\left[\sin\!\left(\frac{i x_i^2}{\pi}\right)\right]^{2m},
\qquad x_i \in [0,\pi],
\end{align}
where we use the standard choice $m=10$.\footnote{\url{https://www.sfu.ca/~ssurjano/michal.html}}

\section{Additional Experiments and Results}
\label{app:add-experiments-results}

\subsection{Samples from a Gaussian Process Prior}
\label{sec:gp_prior_samples}

To further evaluate performance, we generate black-box objectives by sampling functions from an isotropic GP prior with a fixed lengthscale $\ell$ \citep{tang2025nest}. We consider $\ell \in \{0.05,\,0.1,\,0.2\}$ and evaluate $D\in\{50,100\}$ with budgets of 500 and 1{,}000 function evaluations, respectively. We follow an \emph{out-of-model} protocol \citep{stenger2025local}, i.e., the GP hyperparameters used within the BO loop are fit from the queried data (using each method's default training procedure) rather than being set to the data-generating values. As in the main experiments, we refit GP hyperparameters every 10 iterations for all methods except Linear BO.

\begin{figure}[tb!]
  \centering
  \includegraphics[width=0.75\textwidth]{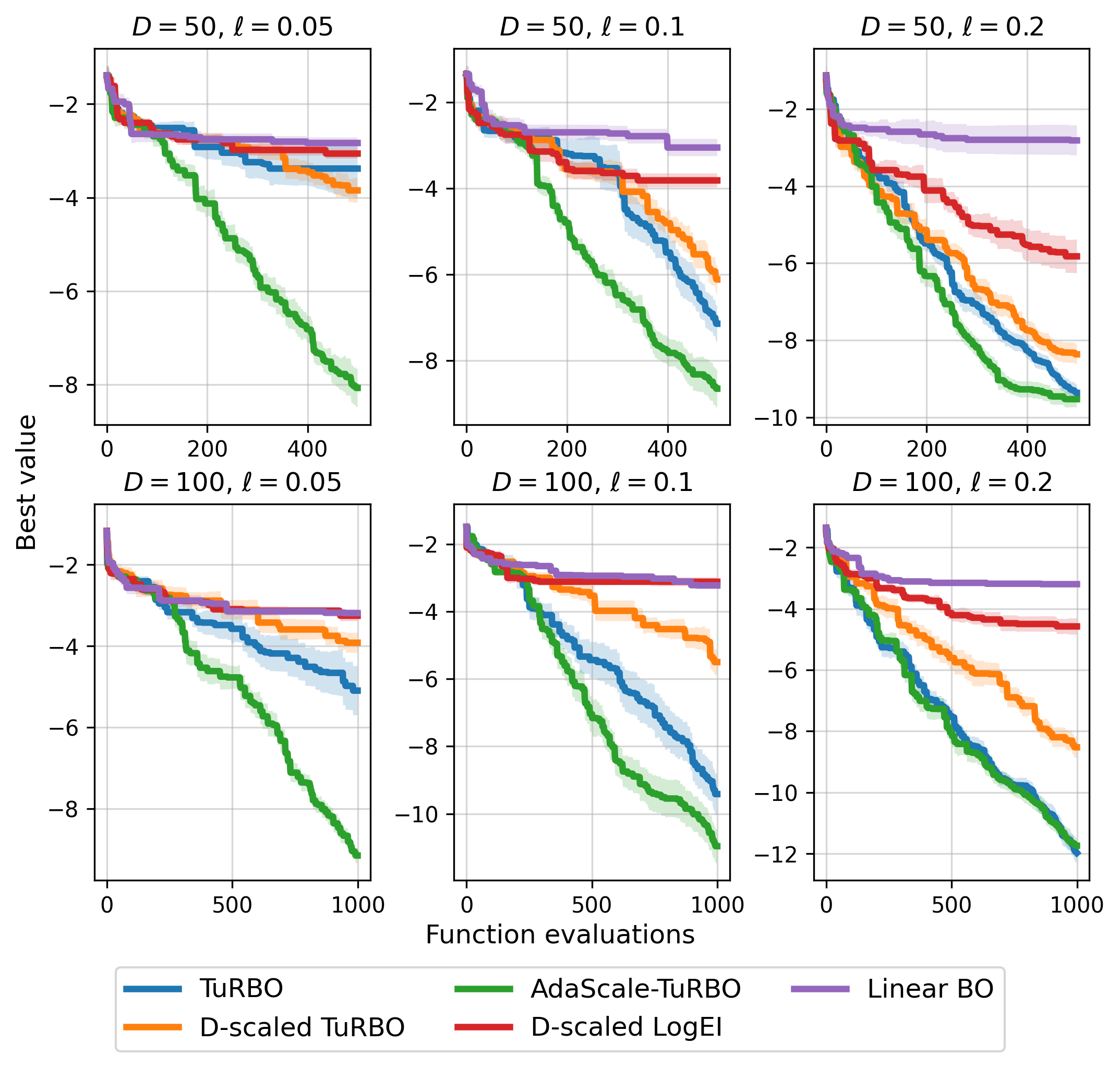}
  \caption{Best observed objective value versus function evaluations on objectives sampled from an isotropic GP prior. Columns correspond to the data-generating lengthscale $\ell \in \{0.05,0.1,0.2\}$. Rows correspond to $D=50$ (top, 500 evaluations) and $D=100$ (bottom, 1{,}000 evaluations). Lines show the median across 10 replicates and shaded bands indicate standard error.}
  \label{fig:GPprior}
\end{figure}

Figure~\ref{fig:GPprior} reports best observed values over the optimization budget. AdaScale-TuRBO achieves the best performance on nearly all settings, and the improvement over TuRBO is most pronounced for smaller $\ell$ (i.e., more rapidly varying functions). This trend is consistent with the intuition that local modeling becomes increasingly beneficial as global smoothness assumptions become harder to satisfy.

\subsection{Ablation of Initial TR Size}
\label{sec:ablation_L0}

We ablate a key TuRBO hyperparameter: the initial TR side length $L_0$. The main text uses $L_0=0.8$ (the default in standard TuRBO implementations). Here we additionally consider $L_0 \in \{0.4,\,0.2\}$ on the $D=50$ synthetic benchmarks from Section~\ref{sec:experiment}, keeping all other TuRBO hyperparameters fixed.

\begin{figure}[tb!]
  \centering
  \includegraphics[width=0.75\textwidth]{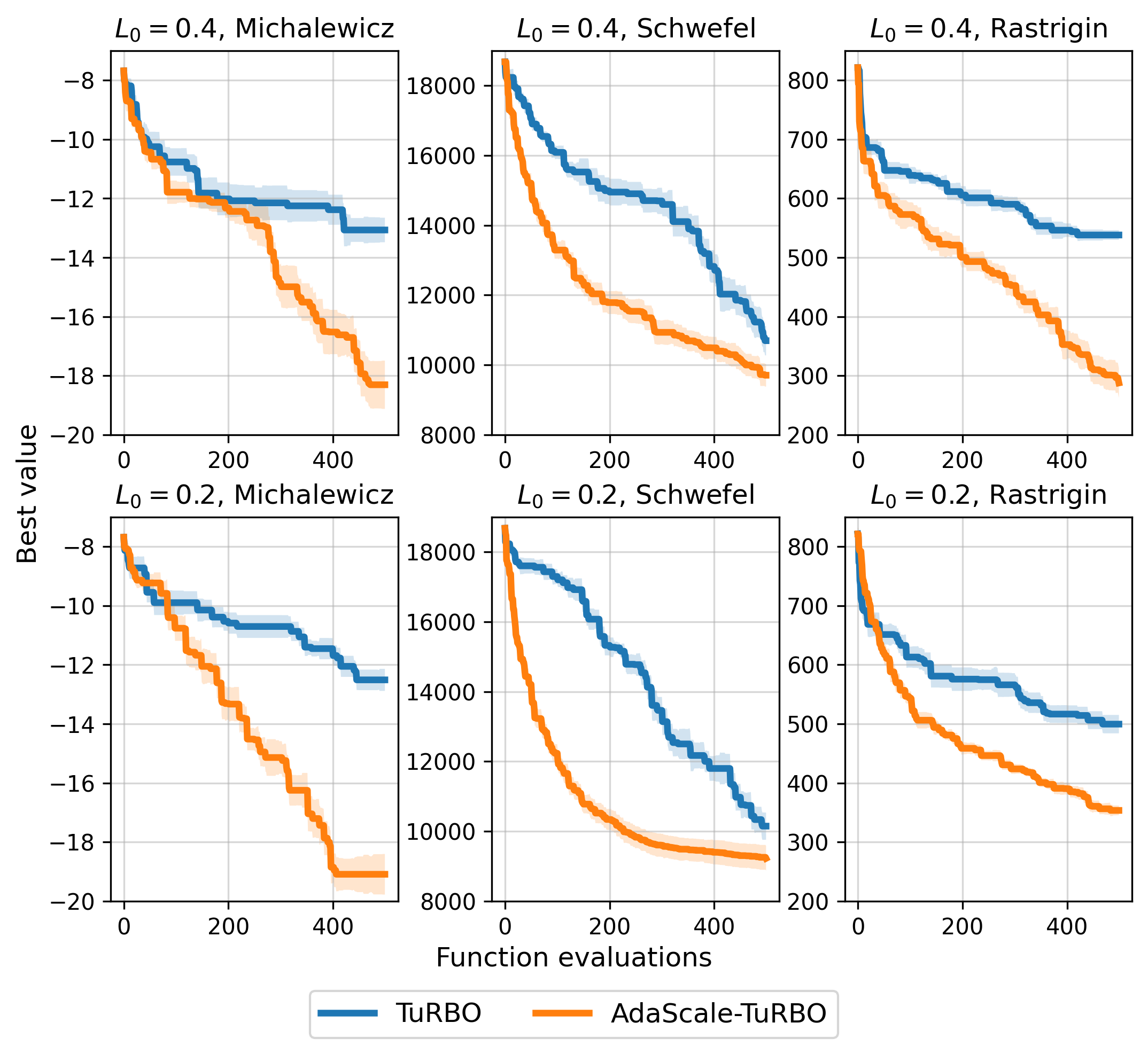}
  \caption{Ablation over the initial trust region side length $L_0$ on $D=50$ synthetic benchmarks (500 evaluations). Top row: $L_0=0.4$. Bottom row: $L_0=0.2$. Lines show the median across 10 replicates and shaded bands indicate standard error.}
  \label{fig:Ablation}
\end{figure}

Figure~\ref{fig:Ablation} shows that AdaScale-TuRBO consistently outperforms TuRBO for both alternative initializations, indicating that the gains are not driven by a particular choice of $L_0$.

\subsection{Best-Found Objective Values}
\label{sec:best_found_value}

\begin{figure}[tb!]
  \centering
  \includegraphics[width=0.95\textwidth]{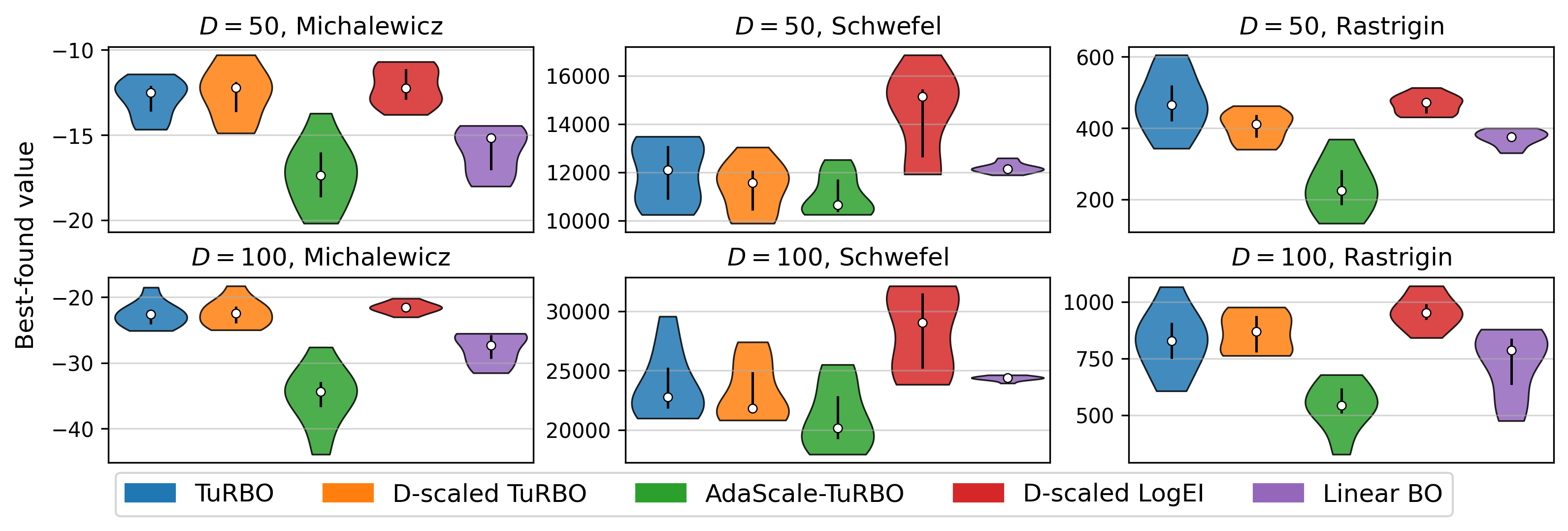}
  \caption{Violin plots of final best-found objective values on the synthetic benchmarks across 10 replicates. Top row: $D=50$ (500 evaluations). Bottom row: $D=100$ (1{,}000 evaluations). White dots indicate medians and thick black bars indicate interquartile ranges (25\%-75\%).}
  \label{fig:Violin_Synthetic}
\end{figure}

\begin{figure}[tb!]
  \centering
  \includegraphics[width=0.95\textwidth]{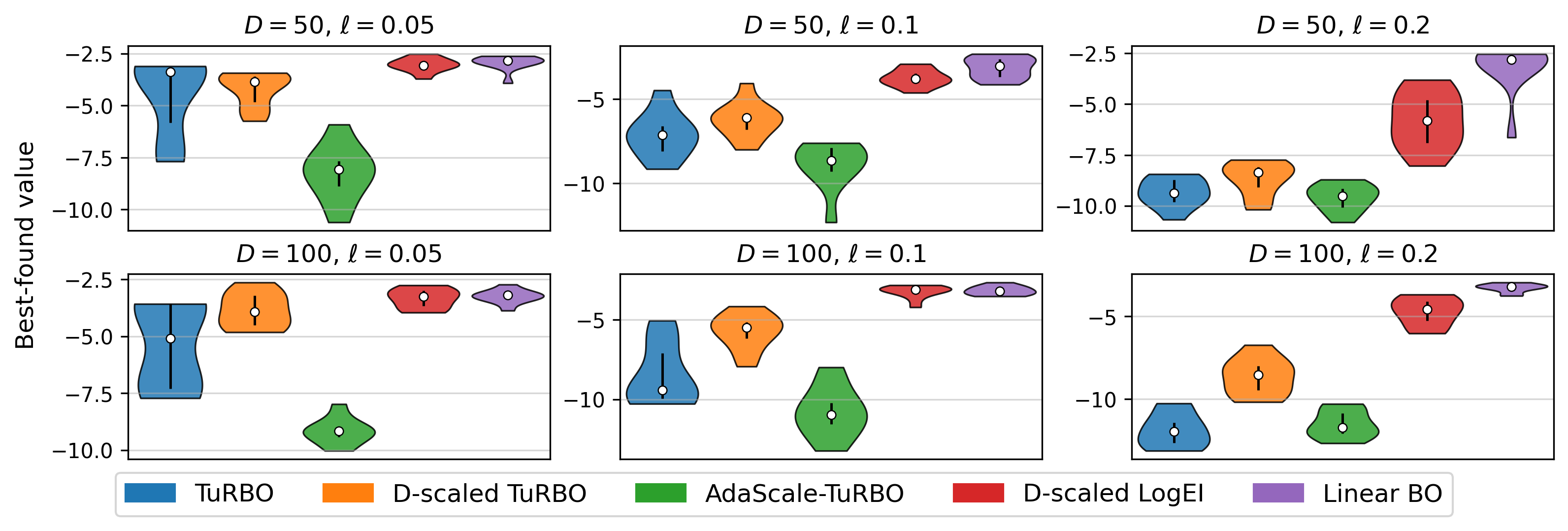}
  \caption{Violin plots of final best-found objective values on GP-prior-sampled objectives across 10 replicates. Columns correspond to the data-generating lengthscale $\ell \in \{0.05,0.1,0.2\}$. Rows correspond to $D=50$ (top, 500 evaluations) and $D=100$ (bottom, 1{,}000 evaluations). White dots indicate medians and thick black bars indicate interquartile ranges (25\%-75\%).}
  \label{fig:Violin_GP}
\end{figure}

\begin{figure}[tb!]
  \centering
  \includegraphics[width=0.8\textwidth]{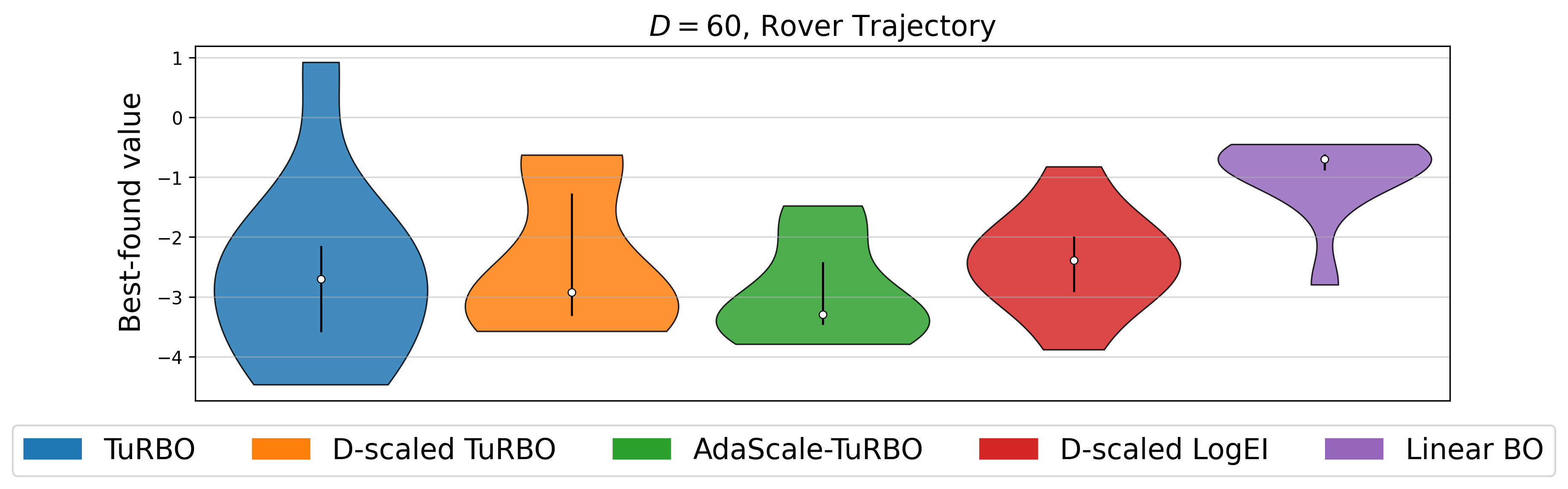}
  \caption{Violin plot of final best-found objective values on the $D=60$ rover trajectory planning benchmark across 10 replicates (1{,}000 evaluations). White dot indicates the median and the thick black bar indicates the interquartile range (25\%-75\%).}
  \label{fig:Violin_Rover}
\end{figure}

We summarize final performance using violin plots of the \emph{best-found} objective value at the end of the evaluation budget across 10 replicates. Figures~\ref{fig:Violin_Synthetic}-\ref{fig:Violin_Rover} report results for (i) synthetic benchmarks, (ii) GP-prior-sampled objectives, and (iii) the rover trajectory planning problem. In all plots, lower values indicate better performance. The white dot denotes the median and the thick black bar denotes the interquartile range (25\%--75\%), with the violin width indicating the empirical density.

\end{document}